\documentclass[sigconf,nonacm,balance=false]{acmart}

\setcopyright{none}
\acmConference[WWW '27]{The ACM Web Conference 2027}{May 10--14, 2027}{Dublin, Ireland}
\acmYear{2027}
\copyrightyear{2027}

\usepackage{graphicx}
\usepackage{amsmath}
\usepackage{multirow}
\usepackage{tabularx}
\usepackage{ragged2e}
\newcolumntype{Y}{>{\RaggedRight\arraybackslash}X}
\usepackage{amsthm}

\usepackage{xspace}

\usepackage[most]{tcolorbox}
\usepackage{CJKutf8}

\usepackage{enumitem}
\usepackage{titletoc}
\usepackage{listings}
\usepackage{pifont}

\tcbuselibrary{listings, breakable}
\usepackage{hyperref}
\usepackage{xurl}
\usepackage{xcolor}
\usepackage{colortbl}
\definecolor{TableHeader}{HTML}{E7EEF4}
\definecolor{TableSection}{HTML}{F2F5F8}
\definecolor{TableHighlight}{HTML}{E9F3EF}
\makeatletter
\newcommand{\TableRowShade}[2]{\noalign{%
    \begingroup\color{#1}%
    \hrule width #2 height \dimexpr\ht\@arstrutbox+\dp\@arstrutbox\relax
    \kern-\dimexpr\ht\@arstrutbox+\dp\@arstrutbox\relax
    \endgroup}}
\makeatother
\usepackage{stfloats}
\usepackage{placeins}

\AtBeginDocument{\renewcommand{\checkmark}{\ding{51}}}

\begin{document}

\raggedbottom

\title{BehaviorMem: Building and Evaluating User Memory from Real-World E-Commerce Behavior Streams}

\author{Sizhe Wang}
\authornote{Equal Contribution.}
\email{wangsz26@m.fudan.edu.cn}
\affiliation{\institution{Fudan University}\city{Shanghai}\country{China}}

\author{Xinghao Fang}
\authornotemark[1]
\email{xhfang26@zju.edu.cn}
\affiliation{\institution{Zhejiang University}\city{Hangzhou}\country{China}}

\author{Rui Qiu}
\authornotemark[1]
\email{qiurui.qiu@taobao.com}
\affiliation{\institution{Taobao \& Tmall Group, Alibaba}\city{Beijing}\country{China}}

\author{Zhenjia Ma}
\email{mazhenjia.mzj@alibaba-inc.com}
\affiliation{\institution{Taobao \& Tmall Group, Alibaba}\city{Beijing}\country{China}}

\author{Dagui Chen}
\authornote{Corresponding Author.}
\email{dagui.cdg@taobao.com}
\affiliation{\institution{Taobao \& Tmall Group, Alibaba}\city{Beijing}\country{China}}

\author{Kun Kuang}
\authornotemark[2]
\email{kunkuang@zju.edu.cn}
\affiliation{\institution{Zhejiang University}\city{Hangzhou}\country{China}}

\author{Hongcheng Guo}
\authornotemark[2]
\email{guohc@fudan.edu.cn}
\affiliation{\institution{Fudan University}\city{Shanghai}\country{China}}

\begin{abstract}
User memory helps personalized systems build on their understanding of users over time. In e-commerce, users convey their preferences and needs through both explicit language and implicit interactions with products. We study memory built from logs of these interactions, examining how it can \emph{remember} concrete evidence and established interests, and use that evidence to \emph{infer} underlying needs and preferences that inform judgments beyond well-observed product domains. Existing memory methods primarily target language-based interactions, and evaluations offer limited real behavioral data and limited separation of these capabilities. This leaves open how to jointly support and separately evaluate remembering and inference from behavioral logs. We formulate the problem of continual behavioral user memory around these joint requirements and develop a dedicated evaluation framework and memory method to address them. \textbf{BehaviorMemBench} is a benchmark built from real e-commerce logs covering 1{,}194 users and 17{,}462 behavioral batches over more than three years, with automatic future-behavior labels and a task-independent memory interface to separately evaluate remembering and inference under a common protocol. \textbf{BehaviorMem} is a memory design that organizes explicit behavioral facts and inferred user characteristics across long-term and short-term scopes in a two-axis state, with evidence-grounded formation, continual maintenance, and query-independent readout. Experiments across two builders and two task readers show that BehaviorMem leads the evaluated agentic memories on remember and infer, and reveal different method rankings across the two capabilities and competitive statistical baselines. Validation on a public dataset further supports its usefulness for downstream tasks based on behavioral histories.

\end{abstract}

\ccsdesc[500]{Information systems~Personalization}
\ccsdesc[500]{Human-centered computing~User models}

\keywords{User memory, behavioral user modeling, personalization, large language models, memory evaluation}

\maketitle

\section{Introduction}
\label{sec:behaviormem-intro}

\begin{figure*}[!tp]
    \centering
    \includegraphics[width=\textwidth]{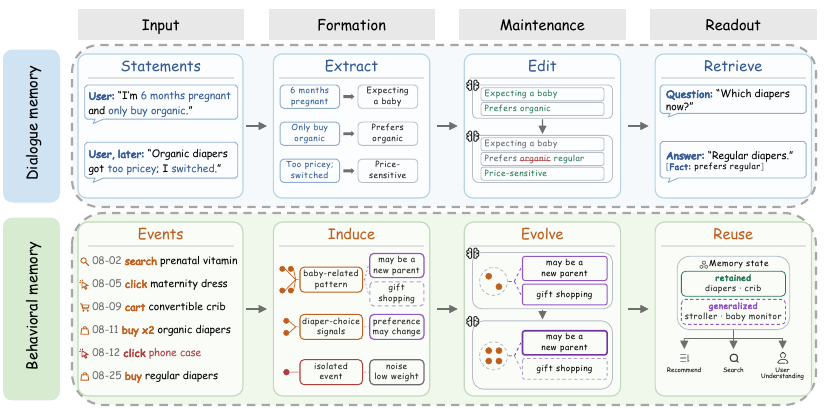}
    \caption{Comparison of dialogue-oriented and behavioral user memory across formation, maintenance, and readout. Dialogue memory preserves and organizes information expressed in interactions. Behavioral memory retains behavioral facts and induces user characteristics from implicit events, maintaining both to support remembering and inference.}
    \label{fig:dialogue-behavior-memory}
\end{figure*}

Long-term personalization in LLM-based systems, such as personal agents, search engines, and recommendation systems, requires persistent user understanding. User memory accumulates and reuses preferences, circumstances, and goals across interactions to support relevant services as users' needs change~\citep{zhong2024memorybank,chhikara2025mem0buildingproductionreadyai}. In e-commerce, this knowledge comes from interactions with products and services, where users convey preferences and shopping needs through both \textbf{explicit natural language} and \textbf{implicit behavior}. Through product inquiries and customer-service conversations, users can directly state their requirements, preferences, and circumstances. By contrast, browsing, favorites, add-to-cart actions, and purchases show what users engage with without directly explaining why. Their underlying preferences and needs must therefore be inferred from product characteristics and patterns across interactions.

Memory construction from explicit natural-language interactions has been extensively studied~\citep{packer2024memgptllmsoperatingsystems,chhikara2025mem0buildingproductionreadyai}. However, implicit behavioral logs provide evidence of preferences and needs that users may not articulate in language. These signals are dispersed across products and visits, requiring observations to be interpreted together over time. Making this evidence useful for later personalization calls for a persistent, updatable source of user information that carries relevant history into future interactions. We therefore focus on constructing \textbf{user memory from behavioral logs} to preserve evidence of users' choices and develop a reusable understanding of the preferences and needs behind them.

Past behavior can inform later personalization. However, behavioral logs record what users did without directly explaining their preferences or needs. Using such evidence therefore calls for memory to \textbf{remember} what users did and \textbf{infer} what their choices imply about them beyond remembering. Remembering keeps concrete choices, established interests, and relevant product or attribute details available for later use. Inference connects recurring choices and shared product characteristics to derive reusable user knowledge that can inform judgments in sparsely observed or unvisited, semantically related product domains. 

Existing memory methods and benchmarks offer \textbf{limited coverage of real, continual behavioral logs}. MemGPT and Mem0 manage dialogue histories, and personalized systems such as MemoryBank and PersonaTree maintain user representations from language-based interactions~\citep{packer2024memgptllmsoperatingsystems,chhikara2025mem0buildingproductionreadyai,zhong2024memorybank,zhao2026insideout}. Figure~\ref{fig:dialogue-behavior-memory} contrasts this setting with implicit e-commerce behavior. Benchmarks such as LoCoMo, LongMemEval, and MemoryAgentBench likewise evaluate language-based interactions~\citep{maharana2024evaluatinglongtermconversationalmemory,wu2025longmemevalbenchmarkingchatassistants,hu2026evaluatingmemoryllmagents}, while behavioral-log benchmarks such as DynamicMem use synthetic trajectories and predefined user states~\citep{xie2026dynamicmemlonghorizonmemorybenchmark}. This motivates memory construction and evaluation grounded in \textbf{real behavioral streams}, with labels automatically derived from future interactions to capture how users' interests unfold in actual use.

A second gap concerns \textbf{joint support and separate evaluation of remember and infer}. Memory methods largely emphasize retaining and retrieving useful information, although some also derive higher-level knowledge: Generative Agents forms reflections, and A-Mem links memories and evolves their contextual representations~\citep{park2023generativeagentsinteractivesimulacra,xu2025amemagenticmemoryllm}. For continual behavioral memory, how to preserve concrete evidence and maintain revisable user inferences together remains underexplored. Evaluation commonly focuses on downstream question answering and the recovery and use of remembered information. LaMP, GISTBench, and StreamProfileBench extend evaluation to personalization or user understanding~\citep{salemi2024lamplargelanguagemodels,fostiropoulos2026gistbenchevaluatingllmuser,wang2026streamprofilebenchbenchmarkfinegraineduser}, but a common assessment separating these capabilities remains limited for continual implicit e-commerce behavior. Addressing this gap calls for a \textbf{task-independent memory interface} and \textbf{separate tests of remember and infer}: whether memory keeps established interests available, and whether it supports future-interest judgments beyond well-observed product domains.

To address these gaps, we introduce \textbf{BehaviorMemBench} for evaluation and \textbf{BehaviorMem} for continual memory construction from e-commerce behavioral logs. BehaviorMemBench covers 1{,}194 users and 17{,}462 behavioral batches over more than three years, using \textbf{automatically derived future-behavior labels} to evaluate how memory supports \textbf{remembering and inference} under a \textbf{common, task-independent protocol}. BehaviorMem addresses the construction gap by \textbf{jointly maintaining behavioral evidence and revisable user understanding}. To \textbf{remember}, its \textbf{two-axis state} organizes explicit facts alongside inferred characteristics across long-term and short-term scopes, keeping past choices and established interests available through \textbf{query-independent readout}. To \textbf{infer}, \textbf{evidence-grounded abstraction} proceeds from behaviors through patterns and themes to user characteristics linked to supporting observations, and \textbf{continual maintenance} revises their content and temporal availability as new evidence arrives, distinguishing enduring tendencies from stage-specific needs.

Our contributions are threefold:
\begin{itemize}
    \item We formulate behavioral user memory around remembering and inference, and introduce \textbf{BehaviorMemBench} with real e-commerce logs, automatic future-behavior labels, and separate evaluation under a common, task-independent protocol.
    \item We propose \textbf{BehaviorMem}, whose two-axis state pairs behavioral facts and inferred characteristics across long-term and short-term scopes, with evidence-grounded abstraction and continual maintenance.
    \item Across two builders and two task readers, BehaviorMem leads the evaluated agentic memories on both recalls. Experiments also reveal different rankings across capabilities, competitive statistical baselines, and complementary memory components, with further validation on downstream tasks.
\end{itemize}

\section{Problem Formulation}
\label{sec:problem-formulation}

\paragraph{\textbf{Behavioral observations.}}
For each user, chronological behavior batches form a stream:
\begin{equation*}
    \mathcal{B}=(B_1,\ldots,B_T),\qquad B_t=\{b_{t,j}\}_{j=1}^{n_t}.
\end{equation*}
Each interaction records a product, its attributes, action type, and timestamp; the user's underlying preferences and needs remain implicit.

\paragraph{\textbf{Causal user-state interface.}}
A memory system updates its state and exposes a query-independent readout:
\begin{equation}
    M_t=U(M_{t-1},B_t),\qquad O_t=\Phi(M_t).
\end{equation}
Both depend only on $B_{\leq t}$, without access to downstream queries, candidate pools, or future labels. The state may combine records, statistical summaries, and inferred characteristics; a common reader evaluates its readout.

\paragraph{\textbf{Joint memory objectives.}}
Memory must \textbf{remember} useful historical evidence and established user information, and \textbf{infer} reusable user knowledge that supports judgments in less-observed or previously unvisited domains. Because abstraction and detail preservation can affect these capabilities differently, we assess them separately.

\paragraph{\textbf{Design questions.}}
BehaviorMemBench operationalizes these objectives through future-behavior tests, while BehaviorMem explores a memory design addressing them.

\section{BehaviorMemBench}
\label{sec:benchmark}

\subsection{Comparison with Existing Benchmarks}
\label{sec:benchmark-comparison}

Table~\ref{tab:benchmark-comparison} compares six properties. For \textbf{data coverage}, existing benchmarks use language-based histories~\citep{maharana2024evaluatinglongtermconversationalmemory,wu2025longmemevalbenchmarkingchatassistants,hu2026evaluatingmemoryllmagents}, synthetic behavior~\citep{xie2026dynamicmemlonghorizonmemorybenchmark}, or real data without continual evaluation~\citep{salemi2024lamplargelanguagemodels,fostiropoulos2026gistbenchevaluatingllmuser}. BehaviorMemBench combines real, continual implicit behavior streams with automatic future-behavior labels. For \textbf{capability evaluation}, DynamicMem and GISTBench are task-independent but do not separately test infer beyond well-observed domains; StreamProfileBench supports both properties using text~\citep{wang2026streamprofilebenchbenchmarkfinegraineduser}. BehaviorMemBench combines these data properties with a task-independent interface and separate remember and infer tests.

\begin{table*}[t]
    \centering
    \caption{Comparison of user-memory and user-profile benchmarks. Infer denotes interest judgments beyond well-observed product domains.}
    \label{tab:benchmark-comparison}
    \small
    \setlength{\tabcolsep}{3pt}
    \begin{tabular*}{\textwidth}{@{\extracolsep{\fill}}llcccccc@{}}
        \toprule
        Type & Benchmark & Continual & \shortstack{Real\\data} & \shortstack{Automatic\\labels} & \shortstack{Implicit\\behavior} & \shortstack{Task-\\independent} & Infer \\
        \midrule
        Memory & LoCoMo & \checkmark & $\times$ & $\times$ & $\times$ & $\times$ & $\times$ \\
        Memory & LongMemEval & \checkmark & $\times$ & $\times$ & $\times$ & $\times$ & $\times$ \\
        Memory & MemoryAgentBench & \checkmark & $\times$ & $\times$ & $\times$ & $\times$ & $\times$ \\
        Memory & DynamicMem & \checkmark & $\times$ & $\times$ & \checkmark & \checkmark & $\times$ \\
     
        Profile & LaMP & $\times$ & \checkmark & \checkmark & $\times$ & $\times$ & $\times$ \\
        Profile & GISTBench & $\times$ & \checkmark & \checkmark & \checkmark & \checkmark & $\times$ \\
        Profile & StreamProfileBench & \checkmark & \checkmark & \checkmark & $\times$ & \checkmark & \checkmark \\
        \midrule
        Ours & BehaviorMemBench & \checkmark & \checkmark & \checkmark & \checkmark & \checkmark & \checkmark \\
        \bottomrule
    \end{tabular*}
\end{table*}

\subsection{Data Collection and Statistics}
\label{sec:benchmark-data}

We construct BehaviorMemBench from Taobao's user behavior logs, spanning February 2023 to June 2026. These logs record browsing, clicks, favorites, add-to-cart actions, and purchases with timestamps and product metadata. Following the visit-merging procedure in Appendix~\ref{app:bench-pipeline-data}, we retain 1{,}194 of 2{,}000 randomly sampled users with at least 50 valid visits. A batch becomes eligible to close after 50 visits or 120 days and closes at the next inter-visit gap exceeding 15 minutes. This combines activity and elapsed-time limits while avoiding a split within a browsing session. The resulting corpus contains 913{,}271 visits, 17{,}462 batches, and 16{,}268 history-to-future transitions; the summary statistics in Appendix~\ref{app:bench-stats} characterize variation across users.

\subsection{Benchmark Framework}
\label{sec:benchmark-framework}

\paragraph{\textbf{Prediction targets and future labels.}}
We use Taobao's item-category taxonomy, which organizes products hierarchically from broad L1 categories through L2 and deeper levels to leaf categories. We select normalized L2 categories as prediction targets: this intermediate level captures recurring interests across items and provides a common evaluation target independent of a specific downstream task. Let $\mathcal{C}$ denote the L2 label space; fixed semantic groups link related L1 domains. At each $t<T$, memory uses only $B_{\leq t}$, while the held-out next batch supplies future labels:
\begin{equation}
    Y_{t+1}=g(B_{t+1})\subseteq\mathcal{C}.
\end{equation}
Positives are derived from favorites, add-to-cart actions, purchases, or qualifying deep browsing. The labeling rules in Appendix~\ref{app:bench-pipeline-eval} exclude shallow clicks alone.

\paragraph{\textbf{Remember and infer pools.}}
Positives are divided by their relation to the observed history. The \textbf{remember pool} tests previously visited categories and unseen categories within well-established L1 domains, measuring how well memory keeps exact categories and established interest domains available. It uses cluster, history, and random distractors. The \textbf{infer pool} tests unseen categories in lightly observed domains (\emph{related}) or unvisited, semantically connected domains (\emph{new}), with history-conditioned and random distractors. Both pools contain three distractors per positive:
\begin{equation}
    \mathcal{C}_t^p=Y_{t+1}^p\cup D_t^p,\quad |D_t^p|=3|Y_{t+1}^p|,\quad p\in\{\mathrm{mem},\mathrm{gen}\}.
\end{equation}
The eligibility and exclusion rules in Appendix~\ref{app:bench-pipeline-eval} define these sets, which are frozen and shared across memory methods.

\paragraph{\textbf{Common readout protocol.}}
A frozen task reader evaluates each pool separately from the same fixed $O_t$. It selects as many categories as the pool contains positives:
\begin{equation}
    n_t^p=|Y_{t+1}^p|,\quad \widehat{Y}_t^p=\operatorname{TaskReader}(O_t,\mathcal{C}_t^p,n_t^p).
\end{equation}
Candidates are deterministically shuffled. Under the common prompt and rendering protocol in Appendix~\ref{app:bench-protocol}, the reader sees neither method identity nor positive labels. It only selects candidates; labels are determined independently from future behavior. 

\paragraph{\textbf{Metrics.}}
Pool recall measures the fraction of future positives recovered, while distractor exposure measures the fraction of an eligible distractor family incorrectly selected:
\begin{equation}
    R_p=\frac{|\widehat{Y}_t^p\cap Y_{t+1}^p|}{|Y_{t+1}^p|},\qquad E_{p,f}=\frac{|\widehat{Y}_t^p\cap D_t^{p,f}|}{|D_t^{p,f}|}.
\end{equation}
Related and new recall evaluate their respective target tiers; higher recall and lower exposure indicate better performance.

\begin{figure}[!tp]
    \centering
    \includegraphics[width=\linewidth]{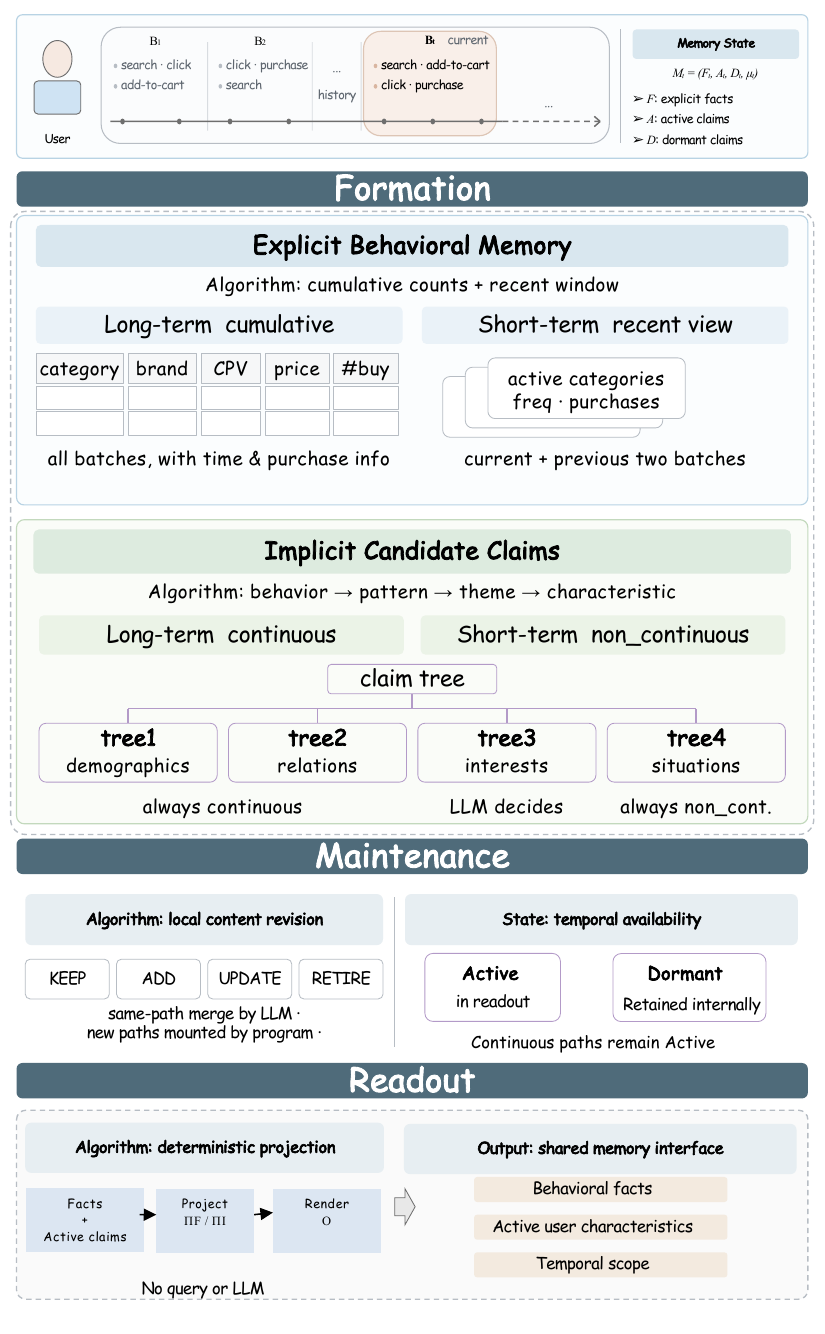}
    \caption{BehaviorMem overview. Programmatic updates preserve explicit behavior, while staged inference and local integration maintain implicit user state. Deterministic projection exposes current memory independently of downstream tasks.}
    \label{fig:behaviormem-overview}
\end{figure}

\section{BehaviorMem}
\label{sec:method}
\label{sec:method-overview}

BehaviorMem targets continual e-commerce behavioral logs, where preferences and needs are implicit in product interactions. Its two-axis design preserves concrete facts to \textbf{remember} past choices and uses evidence-grounded abstraction to \textbf{infer} reusable user characteristics, maintaining long-term tendencies and short-term needs as new behavior arrives. Figure~\ref{fig:behaviormem-overview} summarizes this design.

\subsection{Two-Axis Memory State}
\label{sec:method-state}

The \textbf{explicit/implicit axis} pairs programmatic facts with inferred Claims; the \textbf{long-term/short-term axis} distinguishes accumulated tendencies from recent or stage-specific information:
\begin{equation}
    F_t=(F_t^{\mathrm{L}},F_t^{\mathrm{S}}),\quad M_t=(F_t,A_t,D_t,\mu_t),
    \label{eq:behaviormem-state}
\end{equation}
Here, $F_t^{\mathrm{L}}$ and $F_t^{\mathrm{S}}$ hold historical category and brand aggregates and recent item activity, respectively; $A_t$ and $D_t$ store Active and Dormant Claims on type--domain--attribute paths, and $\mu_t$ tracks path continuity. Claims record user characteristics, confidence, and behavioral references across four types: demographics (T1), social relations (T2), interests (T3), and situational state (T4), with representations specified in Appendix~\ref{app:method-state}.

\subsection{Memory Formation}
\label{sec:method-evolution}
\label{sec:method-formation}

Formation couples programmatic fact accumulation $\mathcal{P}$ with \textbf{evidence-grounded abstraction} $\mathcal{G}_{\theta}$:
\begin{equation}
    F_t=\mathcal{P}(F_{t-1},B_t),\quad Q_t=\mathcal{G}_{\theta}(B_t,\mathcal{R}_{t-1}),
    \label{eq:behaviormem-formation}
\end{equation}
Inference follows \textbf{behavior $\rightarrow$ pattern $\rightarrow$ theme $\rightarrow$ characteristic}, with a further attribute step for T1/T2. Each level uses only its predecessor as evidence. Candidate Claims $Q_t$ retain behavioral references and are placed on semantic paths using registry $\mathcal{R}_{t-1}$, following the procedure in Appendix~\ref{app:method-formation}.

\subsection{Memory Maintenance}
\label{sec:method-maintenance}

Maintenance separates \textbf{local content revision} from \textbf{temporal availability}:
\begin{equation}
    (A_t,D_t,\mu_t)=\mathcal{H}\bigl(A_{t-1},D_{t-1},Q_t,\mu_{t-1}\bigr).
    \label{eq:behaviormem-maintenance}
\end{equation}
Claims are revised locally on matching paths: T1/T2 use deterministic updates, and T3/T4 use source-validated proposals. Retirement requires direct contradiction; missing observations alone never justify deletion. Continuous paths remain Active without renewed observations; stage-specific paths may become Dormant and reactivate with new support. Dormancy hides content from readout while preserving it and its evidence; Appendix~\ref{app:method-maintenance} details the maintenance rules.

\subsection{Memory Readout}
\label{sec:method-readout}

Deterministic projections $\Pi_F$ and $\Pi_I$ expose explicit facts and Active Claims with temporal scope:
\begin{equation}
    O_t=\operatorname{Render}\bigl(\Pi_F(F_t),\Pi_I(A_t,\mu_t)\bigr).
    \label{eq:behaviormem-readout}
\end{equation}
Readout is \textbf{query-independent} and model-free; Dormant content remains internal, and downstream queries, candidates, and labels do not influence construction or projection. The rendering procedure is specified in Appendix~\ref{app:method-readout}.

\section{Experiments}
\label{sec:experiments}

\begin{table*}[!tp]
    \centering
    \caption{BehaviorMemBench results under two task readers ($R$: recall; $E$: distractor exposure).}
    \label{tab:behaviormembench-main}
    \footnotesize
    \setlength{\tabcolsep}{1.8pt}
    \renewcommand{\arraystretch}{1.06}
    \begin{tabular*}{\textwidth}{@{\extracolsep{\fill}}llcccc@{\hspace{7pt}}ccccc@{}}
        \toprule
        & & \multicolumn{4}{c}{\cellcolor{TableHeader}Remember pool} & \multicolumn{5}{c}{\cellcolor{TableHeader}Infer pool} \\
        \cmidrule(lr){3-6}\cmidrule(lr){7-11}
        & & Recall $\uparrow$ & \multicolumn{3}{c}{Exposure $\downarrow$} & \multicolumn{3}{c}{Recall $\uparrow$} & \multicolumn{2}{c}{Exposure $\downarrow$} \\
        \cmidrule(lr){3-3}\cmidrule(lr){4-6}\cmidrule(lr){7-9}\cmidrule(lr){10-11}
        Builder & Method & $R$ & $E_{\mathrm{cluster}}$ & $E_{\mathrm{history}}$ & $E_{\mathrm{random}}$ & $R$ & $R_{\mathrm{related}}$ & $R_{\mathrm{new}}$ & $E_{\mathrm{history}}$ & $E_{\mathrm{random}}$ \\
        \midrule
        \rowcolor{TableSection}
        \multicolumn{11}{c}{\textbf{Task reader: Qwen3-32B}} \\
        \midrule
        \multirow{3}{*}{\shortstack[l]{Non-\\parametric}} & Long Context & \textbf{0.647} & \underline{0.098} & \textbf{0.213} & 0.024 & \textbf{0.426} & \textbf{0.446} & 0.394 & 0.244 & \underline{0.166} \\
         & Frequency Profile & \underline{0.644} & \textbf{0.083} & 0.267 & \textbf{0.015} & \underline{0.424} & \underline{0.435} & \textbf{0.400} & \underline{0.243} & \textbf{0.165} \\
         & Template Profile & 0.593 & 0.116 & \underline{0.245} & \underline{0.020} & 0.417 & \underline{0.435} & \underline{0.397} & \textbf{0.236} & 0.180 \\
        \cmidrule(lr){1-11}
        \multirow{5}{*}{Qwen3-32B} & A-Mem & 0.535 & 0.140 & 0.240 & 0.040 & \underline{0.442} & \underline{0.456} & \textbf{0.416} & \textbf{0.227} & \underline{0.169} \\
         & Mem0 & 0.438 & 0.181 & 0.270 & 0.072 & 0.408 & 0.417 & 0.389 & 0.247 & 0.179 \\
         & LightMem & \underline{0.604} & \underline{0.115} & \textbf{0.220} & \underline{0.032} & 0.422 & 0.441 & \underline{0.395} & 0.240 & 0.172 \\
         & Persona Tree & 0.497 & 0.169 & 0.235 & 0.059 & 0.414 & 0.416 & 0.386 & \underline{0.236} & 0.189 \\
        \TableRowShade{TableHighlight}{\textwidth}
         & \textbf{BehaviorMem (ours)} & \textbf{0.612} & \textbf{0.110} & \underline{0.223} & \textbf{0.024} & \textbf{0.443} & \textbf{0.459} & \textbf{0.416} & \textbf{0.227} & \textbf{0.164} \\
        \cmidrule(lr){1-11}
        \multirow{5}{*}{\shortstack[l]{Qwen3.6-\\35B-A3B}} & A-Mem & 0.543 & 0.124 & 0.271 & 0.037 & 0.424 & 0.437 & 0.404 & 0.240 & 0.166 \\
           & Mem0 & 0.486 & 0.127 & 0.337 & 0.044 & 0.402 & 0.412 & 0.382 & 0.266 & 0.171 \\
           & LightMem & \underline{0.581} & \underline{0.115} & 0.247 & \underline{0.031} & 0.416 & 0.431 & 0.393 & 0.240 & 0.169 \\
           & Persona Tree & 0.543 & 0.134 & \underline{0.239} & 0.048 & \underline{0.436} & \underline{0.451} & \underline{0.412} & \underline{0.237} & \textbf{0.160} \\
        \TableRowShade{TableHighlight}{\textwidth}
           & \textbf{BehaviorMem (ours)} & \textbf{0.611} & \textbf{0.110} & \textbf{0.222} & \textbf{0.022} & \textbf{0.442} & \textbf{0.453} & \textbf{0.418} & \textbf{0.228} & \underline{0.163} \\
        \midrule
        \rowcolor{TableSection}
        \multicolumn{11}{c}{\textbf{Task reader: Qwen3.7-Plus}} \\
        \midrule
        \multirow{3}{*}{\shortstack[l]{Non-\\parametric}} & Long Context & \textbf{0.694} & \underline{0.068} & \textbf{0.226} & 0.017 & \textbf{0.484} & \textbf{0.531} & \textbf{0.419} & \textbf{0.217} & \textbf{0.152} \\
         & Frequency Profile & \underline{0.652} & \textbf{0.057} & 0.317 & \textbf{0.011} & \underline{0.441} & \underline{0.474} & 0.398 & 0.221 & \underline{0.172} \\
         & Template Profile & 0.616 & 0.093 & \underline{0.268} & \underline{0.016} & \underline{0.441} & 0.459 & \underline{0.412} & \underline{0.218} & 0.176 \\
        \cmidrule(lr){1-11}
        \multirow{5}{*}{Qwen3-32B} & A-Mem & 0.565 & 0.101 & 0.292 & \underline{0.025} & 0.462 & 0.472 & 0.425 & 0.221 & 0.166 \\
         & Mem0 & 0.444 & 0.153 & 0.296 & 0.095 & 0.421 & 0.424 & 0.392 & 0.226 & 0.182 \\
         & LightMem & \underline{0.646} & \underline{0.082} & \textbf{0.246} & 0.032 & \underline{0.476} & \textbf{0.499} & \underline{0.430} & 0.217 & \textbf{0.156} \\
         & Persona Tree & 0.522 & 0.129 & \underline{0.262} & 0.063 & 0.429 & 0.432 & 0.410 & \underline{0.215} & 0.177 \\
        \TableRowShade{TableHighlight}{\textwidth}
         & \textbf{BehaviorMem (ours)} & \textbf{0.650} & \textbf{0.070} & 0.274 & \textbf{0.017} & \textbf{0.483} & \underline{0.488} & \textbf{0.447} & \textbf{0.208} & \underline{0.158} \\
        \cmidrule(lr){1-11}
        \multirow{5}{*}{\shortstack[l]{Qwen3.6-\\35B-A3B}} & A-Mem & 0.578 & 0.089 & 0.310 & 0.032 & 0.451 & 0.468 & \underline{0.429} & \underline{0.224} & 0.164 \\
           & Mem0 & 0.502 & 0.095 & 0.392 & 0.035 & 0.421 & 0.446 & 0.386 & 0.260 & \underline{0.161} \\
           & LightMem & \underline{0.635} & \textbf{0.067} & 0.293 & \textbf{0.024} & \underline{0.460} & \textbf{0.484} & 0.413 & 0.230 & \textbf{0.157} \\
           & Persona Tree & 0.575 & 0.101 & \textbf{0.273} & 0.057 & 0.450 & 0.475 & 0.419 & 0.225 & 0.163 \\
        \TableRowShade{TableHighlight}{\textwidth}
           & \textbf{BehaviorMem (ours)} & \textbf{0.646} & \underline{0.072} & \underline{0.276} & \underline{0.029} & \textbf{0.475} & \underline{0.481} & \textbf{0.445} & \textbf{0.203} & 0.162 \\
        \bottomrule
    \end{tabular*}
\end{table*}

\subsection{Experimental Setup}
\label{sec:experimental-setup}

\paragraph{\textbf{Dataset.}}
We use \textbf{BehaviorMemBench} as the primary benchmark to evaluate remembering and inference independently of specific downstream tasks. Its construction and statistics are detailed in Appendix~\ref{app:bench}. To cross-check the effectiveness of memory methods on downstream tasks, we evaluate them on the public \textbf{MemoryCD} benchmark in Section~\ref{sec:public-validation}.

\paragraph{\textbf{Baselines.}}
We include statistical methods to assess what raw behavioral records and simple summaries can support, general-purpose agentic memories to examine whether generic memory mechanisms transfer to behavioral streams, and personal agentic memory to compare designs tailored to user modeling. Implementation details are given in Appendix~\ref{app:bmb-baselines}.
\begin{itemize}
    \item \textbf{Statistical methods.} \textbf{Frequency Profile} ranks categories by engagement and recency, and \textbf{Template Profile} renders behavioral aggregates in a fixed template. \textbf{Long Context} serves as a raw-history control by retaining recent interactions.
    \item \textbf{General-purpose agentic memories.} \textbf{A-Mem}~\citep{xu2025amemagenticmemoryllm} organizes linked memory notes, \textbf{Mem0}~\citep{chhikara2025mem0buildingproductionreadyai} maintains extracted facts, and \textbf{LightMem}~\citep{fang2026lightmemlightweightefficientmemoryaugmented} stores compact long-term entries.
    \item \textbf{Personal agentic memory.} \textbf{Persona Tree}~\citep{zhao2026insideout} maintains a hierarchical persona representation.
\end{itemize}

\paragraph{\textbf{Implementation and evaluation.}}
Methods process history chronologically and expose query-independent readouts. Agentic builders use Qwen3-32B~\citep{yang2025qwen3} or Qwen3.6-35B-A3B~\citep{qwen36_35b_a3b}; Qwen3-32B and Qwen3.7-Plus~\citep{qwen37plus} task readers evaluate the outputs using shared candidate pools and prompts within each reader setting.

\begin{figure*}[!tp]
    \centering
    \includegraphics[width=\textwidth]{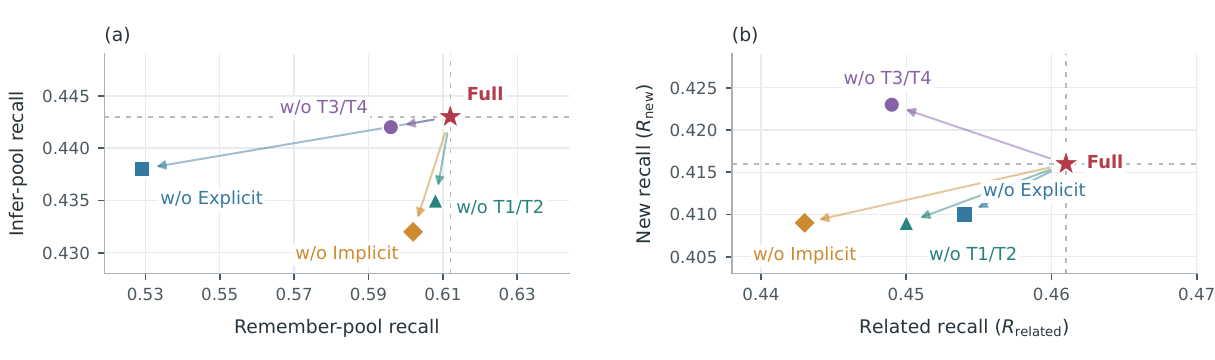}
    \caption{BehaviorMem readout ablations with a Qwen3-32B builder/reader: (a) remember-pool and infer-pool recall; (b) related and new recall. Stars mark Full; arrows indicate masking effects.}
    \Description{Two scatter plots compare Full BehaviorMem with four readout masks. Removing explicit memory mainly reduces remember-pool recall; removing implicit memory lowers infer-pool recall. Masking interest and situational Claims increases new recall while lowering related recall. Stars mark Full, and arrows lead from Full to each mask.}
    \label{fig:readout-ablations}
\end{figure*}

\begin{figure}[!tp]
    \centering
    \includegraphics[width=\linewidth]{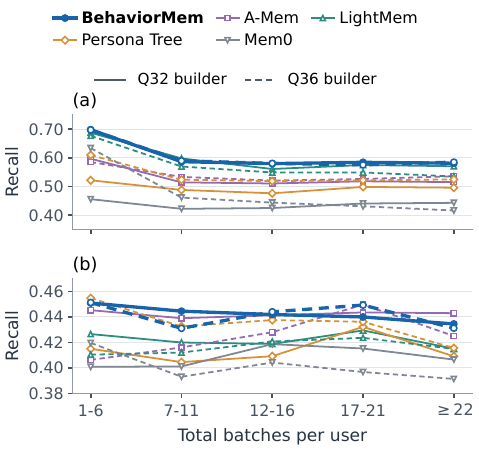}
    \caption{Recall by user activity under a Qwen3-32B task reader, with Qwen3-32B (solid) and Qwen3.6-35B-A3B (dashed) builders: (a) remember-pool recall; (b) infer-pool recall.}
    \Description{Two stacked line charts compare remember-pool recall above and infer-pool recall below across five user-activity groups. Colors and markers identify BehaviorMem, A-Mem, LightMem, Persona Tree, and Mem0. Solid lines use a Qwen3-32B builder and dashed lines use a Qwen3.6-35B-A3B builder. Both panels use a Qwen3-32B task reader, with point estimates only.}
    \label{fig:user-batch-quintiles}
\end{figure}

\begin{figure*}[!tp]
    \centering
    \includegraphics[width=\textwidth]{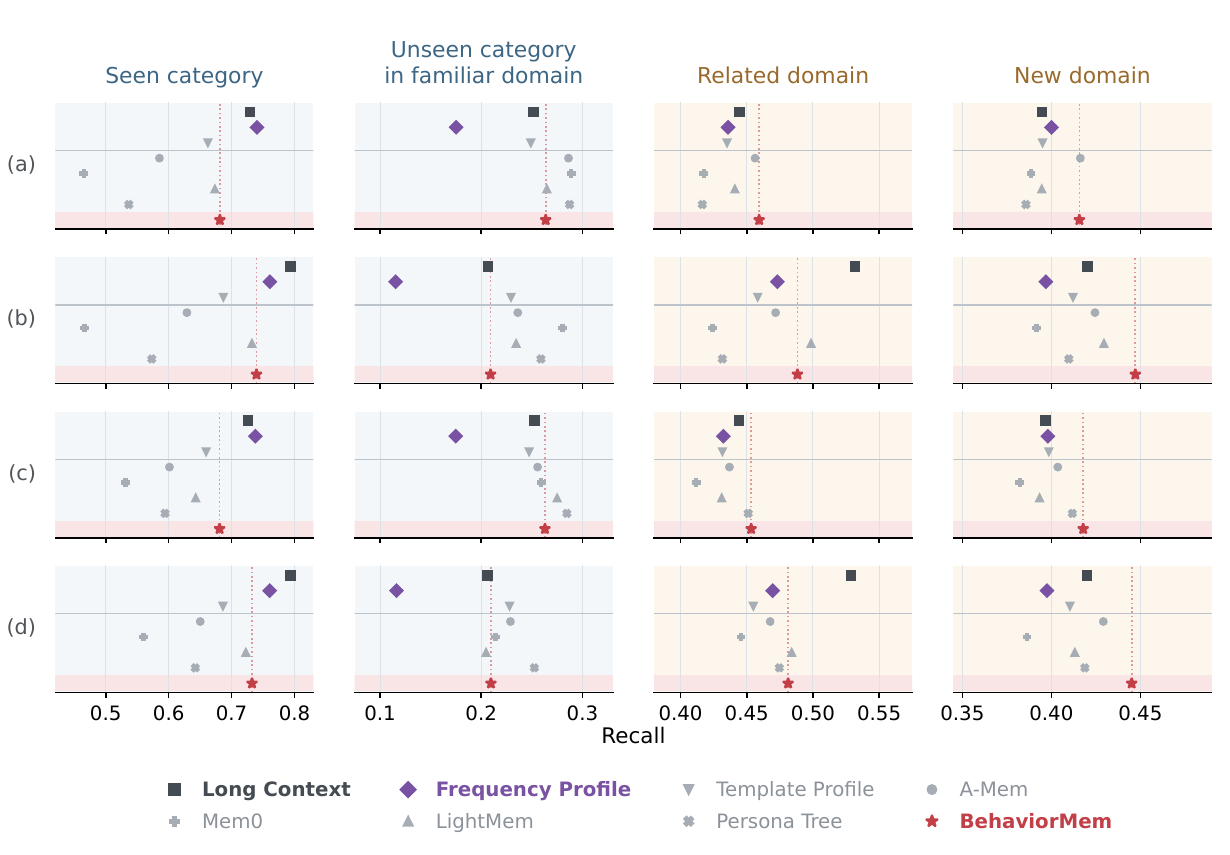}
    \caption{Recall by target familiarity. Builder/reader: (a) Qwen3-32B/Qwen3-32B; (b) Qwen3-32B/Qwen3.7-Plus; (c) Qwen3.6-35B-A3B/Qwen3-32B; (d) Qwen3.6-35B-A3B/Qwen3.7-Plus. Dotted lines mark BehaviorMem.}
    \Description{Four rows of horizontal dot plots compare eight memory methods across seen categories, unseen categories in familiar domains, related domains, and new domains. Rows a and b use a Qwen3-32B builder, and rows c and d use a Qwen3.6-35B-A3B builder. Rows a and c use a Qwen3-32B reader; rows b and d use Qwen3.7-Plus. Long Context, Frequency Profile, and BehaviorMem are highlighted; five other methods use gray points with distinct shapes.}
    \label{fig:target-familiarity}
    \label{fig:target-familiarity-q36}
\end{figure*}

\subsection{Overall Results}
\label{sec:main-results}

Table~\ref{tab:behaviormembench-main} groups results by task reader and memory builder.

\paragraph{\textbf{Remembering and inference favor different memory representations.}}
The two pools reveal different method rankings. With the Qwen3-32B builder and task reader, LightMem achieves higher remember-pool recall than A-Mem, whereas A-Mem achieves higher infer-pool recall. Thus, remembering established interests more effectively does not necessarily translate into better anticipation of future categories beyond the observed history. Evaluating both pools exposes differences that either recall measure alone would miss.

\paragraph{\textbf{BehaviorMem leads agentic memories in remembering and inference.}}
Although individual agentic baselines approach BehaviorMem's performance in particular configurations, BehaviorMem consistently achieves the highest remember-pool and infer-pool recall across all four builder--reader configurations. Its key advantage is this joint performance: the same maintained memory keeps established interests available while supporting interest judgments beyond well-observed domains. This combination makes BehaviorMem well suited to task-independent personalization, where accumulated user knowledge must remain useful across changing contexts and prediction targets.

\paragraph{\textbf{Raw history and statistical memories remain strong controls.}}
Long Context and Frequency Profile remain competitive, particularly for remembering established interests. Their performance highlights the continued value of raw behavioral history and simple statistical summaries as strong baselines for behavioral user memory.

\paragraph{\textbf{Results across builders and readers.}}
Recall patterns vary across builders, while BehaviorMem's recalls remain close within each reader setting. BehaviorMem leads the evaluated agentic memories on both recall measures under both task readers.

\subsection{Ablation}
\label{sec:ablation}
We examine how different memory components support remembering and inference. Figure~\ref{fig:readout-ablations} compares the Full readout with variants that separately mask explicit behavioral facts, implicit user characteristics, T1/T2 Claims, or T3/T4 Claims. Complete results appear in Table~\ref{tab:ablation-main} in Appendix~\ref{app:exp-behaviormembench}.

\paragraph{\textbf{Explicit memory and T3/T4 Claims support remembering.}}
Removing explicit memory reduces remember-pool recall from 0.612 to 0.529, the largest decline among the masks. Removing implicit memory yields a smaller decline to 0.602; masking T3/T4 also has a greater effect than masking T1/T2. Explicit evidence is therefore the main basis for remembering in this readout, with inferred interests and situational needs providing complementary support.

\paragraph{\textbf{Implicit memory supports inference, particularly through T1/T2.}}
Removing implicit memory reduces infer-pool recall from 0.443 to 0.432; among Claim-type masks, removing T1/T2 has a greater effect than removing T3/T4. This suggests that demographic and social characteristics provide useful transferable cues in this configuration. The contribution is not uniform across tiers: masking T3/T4 raises new recall from 0.416 to 0.423, despite reducing remember-pool recall. The two objectives and their subgroups therefore expose different effects of the same memory content.

\subsection{Discussion}
\label{sec:discussion}
\label{sec:memory-strategy-comparison}
\textbf{Overall averages can hide differences across users and targets.} We therefore examine performance by breakdown users:

\paragraph{\textbf{User activity.}}
\label{sec:user-history-analysis}
Figure~\ref{fig:user-batch-quintiles} groups users by total batch count, a proxy for activity, and compares both builders under a Qwen3-32B task reader. For the Qwen3-32B builder, remember-pool recall is highest in the least-active group across all five agentic methods, while infer-pool recall does not improve consistently with activity. BehaviorMem's remember-pool recall exceeds A-Mem's in every group; their infer-pool recalls are close, with A-Mem slightly ahead in the two most-active groups. Activity thus reveals variation in relative support for inference that overall rankings can obscure.

\paragraph{\textbf{Target familiarity.}}
\label{sec:target-familiarity-analysis}
Figure~\ref{fig:target-familiarity} compares both builders under both readers. Frequency Profile is competitive on seen categories but trails other methods on unseen categories within familiar domains. Its strength in recovering repeated interests does not extend uniformly to new categories. Under the Plus reader, Long Context leads on related targets, whereas BehaviorMem leads on new targets. These rankings suggest complementary value in preserving behavioral detail and forming reusable user characteristics.

\textbf{Downstream use of memory depends on its capabilities and is constrained by readout token length.} Besides predictive performance, we also examine the memory input required by task readers. BehaviorMem averages 2{,}018 tokens with Qwen3-32B and 1{,}850 with Qwen3.6-35B-A3B. Long Context averages 7{,}338 tokens, and LightMem averages 7{,}084 and 6{,}812 for the two builders. BehaviorMem thus uses over 70\% fewer memory tokens than either method. Together with its leading recalls among agentic memories, these results show that BehaviorMem supports both capabilities through compact readouts, reducing the amount of memory text supplied to downstream task readers.

\begin{table*}[!tp]
    \centering
    \caption{MemoryCD results under a Qwen3.6-35B-A3B task reader, with both memory builders.}
    \label{tab:memorycd-builders}
    \label{tab:memorycd-compact}
    \label{tab:memorycd-builder-q32}
    \label{tab:memorycd-builder-q36}
    \label{tab:memorycd-reader-q36}
    \footnotesize
    \setlength{\tabcolsep}{2.0pt}
    \renewcommand{\arraystretch}{0.98}
    \begin{tabular*}{0.96\textwidth}{@{\extracolsep{\fill}}lcccc@{\hspace{6pt}}cccc@{}}
        \toprule
        \rowcolor{TableHeader}
        Memory & \multicolumn{4}{c}{\textbf{Single-domain}} & \multicolumn{4}{c}{\textbf{Cross-domain}} \\
        \cmidrule(lr){2-5}\cmidrule(lr){6-9}
        & HR@10 $\uparrow$ & N@10 $\uparrow$ & MAE $\downarrow$ & RMSE $\downarrow$ & HR@10 $\uparrow$ & N@10 $\uparrow$ & MAE $\downarrow$ & RMSE $\downarrow$ \\
        \midrule
        \rowcolor{TableSection}
        \multicolumn{9}{l}{\emph{Non-parametric memories}} \\
        Long Context & \textbf{0.567} & \textbf{0.309} & \textbf{0.485} & \textbf{0.880} & 0.483 & 0.218 & \textbf{0.512} & \textbf{0.843} \\
        Frequency Profile & 0.526 & 0.253 & 0.893 & 1.279 & \underline{0.497} & \underline{0.235} & 1.634 & 1.832 \\
        Template Profile & \underline{0.556} & \underline{0.295} & \underline{0.803} & \underline{1.146} & \textbf{0.512} & \textbf{0.248} & \underline{1.463} & \underline{1.659} \\
        \cmidrule(lr){1-9}
        \rowcolor{TableSection}
        \multicolumn{9}{l}{\emph{Builder: Qwen3-32B}} \\
        Mem0 & 0.515 & 0.255 & 0.768 & 1.162 & 0.476 & 0.204 & 1.198 & 1.603 \\
        A-Mem & \underline{0.604} & \underline{0.341} & 0.726 & 1.134 & 0.524 & 0.252 & 1.047 & 1.311 \\
        Persona Tree & 0.590 & 0.331 & 0.719 & 1.109 & \underline{0.548} & \underline{0.254} & 0.938 & 1.237 \\
        LightMem & 0.576 & 0.335 & \textbf{0.586} & \textbf{0.989} & 0.508 & 0.234 & \textbf{0.678} & \textbf{1.001} \\
        \TableRowShade{TableHighlight}{0.96\textwidth}
        \textbf{BehaviorMem} & \textbf{0.616} & \textbf{0.351} & \underline{0.650} & \underline{1.046} & \textbf{0.563} & \textbf{0.280} & \underline{0.756} & \underline{1.039} \\
        \cmidrule(lr){1-9}
        \rowcolor{TableSection}
        \multicolumn{9}{l}{\emph{Builder: Qwen3.6-35B-A3B}} \\
        Mem0 & 0.517 & 0.257 & 0.785 & 1.170 & 0.470 & 0.200 & 1.224 & 1.609 \\
        A-Mem & \underline{0.586} & \underline{0.332} & 0.778 & 1.172 & \underline{0.537} & \underline{0.253} & 0.902 & 1.180 \\
        Persona Tree & 0.582 & 0.326 & 0.800 & 1.195 & 0.511 & 0.240 & 0.923 & 1.214 \\
        LightMem & 0.565 & 0.318 & \textbf{0.629} & \textbf{1.016} & 0.521 & 0.234 & \textbf{0.778} & \textbf{1.067} \\
        \TableRowShade{TableHighlight}{0.96\textwidth}
        \textbf{BehaviorMem} & \textbf{0.610} & \textbf{0.355} & \underline{0.731} & \underline{1.100} & \textbf{0.544} & \textbf{0.271} & \underline{0.820} & \underline{1.091} \\
        \bottomrule
    \end{tabular*}
\end{table*}

\subsection{Validation on Public Dataset}
\label{sec:public-validation}

\paragraph{\textbf{Setup.}}
We further evaluate on \textbf{MemoryCD}~\citep{zhang2026memorycd}, built from Amazon Reviews 2023. Single-domain evaluation covers Beauty, Books, Electronics, and Home and Kitchen; cross-domain evaluation builds memory from the first three domains and tests on Home and Kitchen for a 323-user cohort. We use the baseline families in Section~\ref{sec:experimental-setup}. Table~\ref{tab:memorycd-builders} reports both memory builders under a Qwen3.6-35B-A3B reader on four metrics. Implementation details and evaluation conventions appear in Appendix~\ref{app:exp-memorycd}.

\paragraph{\textbf{Item ranking.}}
BehaviorMem achieves the highest HR@10 and NDCG@10 across both builders and domain settings under the Qwen3.6-35B-A3B reader, outperforming both agentic memories and non-parametric baselines. For the Qwen3-32B builder, its cross-domain scores reach 0.563 and 0.280, respectively, compared with 0.548 and 0.254 for Persona Tree, the strongest competitor in this setting. The gains in both metrics indicate improvements in placing relevant items within the top ten and ranking them highly. The cross-domain results further support the usefulness of the maintained user representation when the target domain differs from the domains used for memory construction.

\paragraph{\textbf{Rating prediction.}}
Under this reader, BehaviorMem ranks second among agentic memories on both MAE and RMSE across both builders and domain settings, consistently improving on A-Mem, Mem0, and Persona Tree. LightMem has the lowest errors among agentic memories, while Long Context remains best overall. Ranking requires distinguishing preferred items within a candidate set, whereas rating prediction additionally requires matching the user's numerical rating scale. The differing method rankings show that accurate preference ordering and rating calibration place complementary demands on user memory, motivating evaluation across both downstream tasks.

\section{Conclusion}
\label{sec:conclusion}

We study user memory from real, continual e-commerce behavioral logs, where preserving observed choices provides the foundation for inferring the preferences and needs behind them. This setting calls for joint support for \emph{remembering} and \emph{inference}: keeping established user information available and making behavioral evidence useful beyond well-observed product domains. We introduce \textbf{BehaviorMemBench} to assess these capabilities separately through automatically derived future-behavior labels and a common, task-independent protocol. We also propose \textbf{BehaviorMem}, which jointly maintains behavioral facts and revisable user characteristics through a two-axis state, evidence-grounded abstraction, continual maintenance, and query-independent readout. Across two builders and two task readers, BehaviorMem leads evaluated agentic memories on both recalls. Ablations show complementary roles for observed facts and inferred characteristics, while statistical baselines remain competitive. Validation on a public dataset further supports its downstream utility.

\section{Related Work}
\label{sec:related-work}

\paragraph{\textbf{Memory for LLM Agents.}}
Memory enables LLM agents to move beyond stateless interactions by retaining and reusing past information to support reasoning and adaptation.
Early studies represent memory as natural-language records of experience and manage them across hierarchical tiers \citep{park2023generativeagentsinteractivesimulacra,packer2024memgptllmsoperatingsystems,kang2025memoryos}.
Later research distills interactions into structured entries and exposes relations among them \citep{chhikara2025mem0buildingproductionreadyai,xu2025amemagenticmemoryllm,xia2026memora,hu2026evermemos}.
Among recent efforts, entries are consolidated by shared topic and linked through multiple types of relations \citep{liu2026simplememefficientlifelongmemory,fang2026lightmemlightweightefficientmemoryaugmented,jiang2026magmamultigraphbasedagentic,tan2025reflectivememory,su2026temporalsemanticmemory,sun2026preferenceaware}.
Together, these advances give agents symbolic structure and scalable access, all grounded in explicit interactions such as conversations.
BehaviorMem, in contrast, builds memory from implicit behavioral records such as purchases and clicks, which are noisier and less self-descriptive.

\paragraph{\textbf{User Modeling in E-commerce.}}
User modeling supports personalized search and recommendation in e-commerce by capturing shoppers' preferences from their behavioral histories.
Traditional methods encode behavior sequences into latent user representations through attention-based interest modeling over historical actions \citep{zhou2018deepnetworkclickthroughrate,zhou2018deepevolutionnetworkclickthrough,li2019multiinterestnetworkdynamicrouting,hidasi2016session,kang2018selfattentive,sun2019bert4rec,cen2020controllable,yu2019adaptive,pi2020searchbased,zheng2022disentangling}.
Recent LLM-based approaches instead express user preferences in language and preserve this understanding as user memory for subsequent interactions.
Some approaches construct user memory by distilling diverse behavioral patterns into natural-language descriptions of users' multiple interests \citep{shi2025personaxrecommendationagentoriented,choi2026hierarchicalmultipersonainductionuser,sabouri2025towards,ramos2024transparent,wang2025lettingo,wongso2025genup}, while others revise it through interaction feedback and collaborative signals across users \citep{zhang2023agentcfcollaborativelearningautonomous,chen2026memreccollaborativememoryaugmentedagentic,bang2025llmbased,liao2026atom}.
Beyond this, BehaviorMem combines structured organization of user inferences and their supporting events with temporal management of memory revision and current availability.

\paragraph{\textbf{Evaluation of User Memory.}}
User memory evaluation examines what memory retains and how well it supports an understanding of the user.
Existing benchmarks test recall and reasoning over long-term interactions \citep{maharana2024evaluatinglongtermconversationalmemory,wu2025longmemevalbenchmarkingchatassistants,tan2025membench,bian2026realmem} and memory processing under incremental information streams \citep{hu2026evaluatingmemoryllmagents}.
User-focused evaluations assess evolving profiles inferred from synthetic activity logs \citep{xie2026dynamicmemlonghorizonmemorybenchmark,jiang2025know}, personalized task performance \citep{salemi2024lamplargelanguagemodels}, and the alignment of inferred interests with user engagement \citep{fostiropoulos2026gistbenchevaluatingllmuser}.
StreamProfileBench uses future user-authored content to label interests in streaming evaluation, distinguishing historically seen from newly appearing interests \citep{wang2026streamprofilebenchbenchmarkfinegraineduser}.
Building on these directions, BehaviorMemBench combines continual memory construction from real-world implicit behaviors with automatically derived labels from future transaction categories.
Its history-aware candidate pools provide separate operational measures of remembering and interest inference.

\section*{Reproducibility and Ethics Statement}
The use of behavioral data was authorized by the platform and underwent privacy and ethics review. BehaviorMemBench primarily contains item information and interaction records, without direct user profile information, and will be publicly released after data processing and review. Section~\ref{sec:experimental-setup} specifies baselines and model assignments; Appendices~\ref{app:bench-protocol}, \ref{app:method}, \ref{app:exp-behaviormembench}, and~\ref{app:exp-memorycd} document construction details, abridged prompts, baseline adaptations, and evaluation protocols.

\section*{Acknowledgment}
This work was supported by Alibaba Group through Alibaba Innovative Research Program.

\bibliographystyle{ACM-Reference-Format}
\bibliography{custom}

\appendix
\setlength{\emergencystretch}{2em}

\section{BehaviorMemBench Details}
\label{app:bench}

\subsection{Data Collection Pipeline}
\label{app:bench-pipeline}
Data processing produces observations $B_1,\dots,B_T$ independently of evaluation targets. Evaluation construction then labels each history-to-future transition; memory construction and readout never access its candidates.

\subsubsection{Part I: Data processing}
\label{app:bench-pipeline-data}
Taobao interaction logs span 2023-02-03 to 2026-06-23. Of 2{,}000 randomly sampled users, 1{,}194 have at least 50 valid visits. Events contain timestamps, actions, item text, category paths, brand, shop, price, and product attributes.

Consecutive same-item events with gaps of at most 30 minutes merge into one visit; $A\rightarrow B\rightarrow A$ remains three visits. Metadata comes from the first event. Four counters retain entry clicks, detailed browsing (14 action types), favorites/cart actions (6 types, including buy-now intent), and completed purchases. Cart removals, favorite cancellations, and refunds are dropped rather than subtracted.

A batch becomes eligible at 50 visits or 120 days from its first visit and closes at the next inter-visit gap exceeding 15 minutes. This balances sparse and active users without splitting an ongoing session.

\subsubsection{Part II: Evaluation construction}
\label{app:bench-pipeline-eval}
Prediction uses normalized L2 categories within the L1--L2--leaf taxonomy; placeholder and unsuitable categories are removed. Next-batch favorites, cart actions, purchases, or qualifying deep browsing supply positives; shallow clicks alone do not.

\textbf{Pool assignment} depends only on preceding history. Remember positives are seen categories or unseen categories within frequently visited L1 domains. Infer positives are unseen categories in lightly observed domains (related), or in unvisited domains linked to history by a fixed semantic group (new). Unsupported categories are excluded.

\textbf{Distractors} match the pool's historical relation. Remember uses unseen siblings under positive L1 domains (cluster), seen categories under other domains (history), and eligible random fill. Infer uses unseen siblings anchored to historical interests in lightly observed domains (history-conditioned) and random fill; new-domain distractors have unvisited L1 domains in semantic groups represented in history.

All distractors exclude every category in the next batch and are deduplicated; infer distractors also exclude seen categories. Each retained positive requires three legal distractors. Infeasible groups are trimmed from their deterministic tail until this ratio holds. Pools are constructed once with deterministic randomness, shuffled, and frozen across methods and readers.

\subsection{Benchmark Statistics}
\label{app:bench-stats}
The corpus contains 913{,}271 visits, 17{,}462 batches, and 16{,}268 transitions, spanning 2{,}119 L2 categories in 161 L1 domains; the normalized taxonomy contains 4{,}667 L2 categories. Median visits per user, batches per user, visits per batch, and batch duration are 736, 14, 52, and 14 days. Batch duration is available for 17{,}440 batches. Activity varies substantially: 101 users have at most three batches and 501 have at least 16. Full distributions are in the repository.

\subsection{Readout and Task-Reader Protocol}
\label{app:bench-protocol}
Methods in each reader block share candidate pools, prompts, and decoding settings. Fixed readouts fill common description/tree slots, with fixed placeholders for unused slots. Remember and infer use separate calls on the same state. Candidates show zero-based indices, L2 names, and shared short annotations; method identity, labels, and L1 parents are hidden.

Empty positive pools are skipped. The following task-reader template is translated from Chinese; the same selection format is used for both pools.

\begin{tcolorbox}[breakable, colback=black!2, colframe=black!25, boxrule=0.3pt, arc=1mm, left=4pt, right=4pt, top=3pt, bottom=3pt, before skip=4pt, after skip=4pt]
\footnotesize\ttfamily\RaggedRight
You are a user-profile analysis system. Based on the user profile, select the categories that best match the user's interests. Use only the profile description and profile tree.\\[0.3em]
\#\#\# Profile description: \{description\}\\
\#\#\# Profile tree: \{tree\}\\
\#\#\# Candidate pool B (\{pool\_size\} candidates)\\
\{pool\}\\
select \{n\}\\[0.3em]
**Output format**:\\
\#\# Selection\\
Pool B: 0,2,4\\
- Indices start from 0\\
- Select \{n\} from pool B\\
- Separate the indices with commas; do not explain
\end{tcolorbox}

The nine metrics follow Section~\ref{sec:benchmark-framework}. Exposure normalizes selected distractors by family size. Each metric skips empty denominators, averages valid transitions within users and then across users, excluding users without valid transitions. Cohorts are defined per reader; each builder--reader configuration uses common valid transitions.

\section{BehaviorMem Implementation Details}
\label{app:method}

\subsection{Per-Batch Execution}
\label{app:method-exec}
F1 renders behavior; F2 derives candidate Claims; F3 places them; F4 updates state; F5 projects readout. T1/T2 use closed slots and T3/T4 use open Claims, merging into one tree before F5. An independent append-only visit ledger records F1 facts and never enters the Claim tree or upstream model prompts.

\subsection{State Representation}
\label{app:method-state}
Explicit long-term memory aggregates categories, brands, attributes, and engagement with recency decay; short-term memory retains item activity from the latest three batches. Implicit types are demographics (T1), social relations (T2), interests (T3), and situational state (T4), organized by type--domain--attribute paths.

T1/T2 use fixed registry paths with at most one current value; T3/T4 reuse or create paths and allow multiple Claims. Products, brands, values, and assertions cannot name paths. Claims store text, confidence from 1 to 5, and first-seen/last-confirmed endpoints with dates, key facts, and visit references. T1/T2 are continuous and bypass Dormancy; T4 is non-continuous. Continuous T3 Claims represent longer-term tendencies; non-continuous T3/T4 Claims represent stage-specific needs. Dormant Claims remain stored but hidden from readout.

\subsection{Formation: The Evidence-Distance Ladder}
\label{app:method-formation}
Formation follows behavior $\rightarrow$ recurring commonality $\rightarrow$ attended theme $\rightarrow$ reusable characteristic $\rightarrow$ demographic/social attribute. A shared call derives the first two abstraction levels. Two arms then use the same themes: T3/T4 stop at a free-text characteristic; T1/T2 continue to a value copied from a closed candidate set. Each level uses only the immediately preceding level as evidence. Stored terminals retain behavioral references, not a materialized parent chain.

The following abridged formation prompt is translated from Chinese; D0--D3 denote the four abstraction levels above.

\begin{tcolorbox}[breakable, colback=black!2, colframe=black!25, boxrule=0.3pt, arc=1mm, left=4pt, right=4pt, top=3pt, bottom=3pt, before skip=4pt, after skip=4pt]
\footnotesize\ttfamily\RaggedRight
        You are an e-commerce behavioral inference system.\\[0.3em]
        \#\# Shared stage\\
        Input: \{behavior log\}\\
        Derive D0 behavioral commonalities from observed events only, preserving Visit references and merging repeated behavior toward the same object or theme. Derive D1 objects or themes of attention from D0 only.\\[0.3em]
        \#\# T1/T2 arm\\
        Input: \{D1\}, \{closed-set registry\}\\
        Derive D2 from D1 only, then copy one allowed D3 value verbatim.\\[0.3em]
        \#\# T3/T4 arm\\
        Input: \{D1\}, \{open registry\}\\
        Derive a reusable free-text D2 characteristic from D1 only. Do not introduce behavior absent from the preceding level.
\end{tcolorbox}

T1/T2 already have registry paths. The following abridged placement prompt applies only to T3/T4.

\begin{tcolorbox}[breakable, colback=black!2, colframe=black!25, boxrule=0.3pt, arc=1mm, left=4pt, right=4pt, top=3pt, bottom=3pt, before skip=4pt, after skip=4pt]
\footnotesize\ttfamily\RaggedRight
You are a user-memory placement system.\\[0.3em]
Input: \{current T3/T4 registry\}, \{T3/T4 candidate Claims\}\\[0.3em]
Perform placement only. For every candidate, select exactly one type, domain, and attribute, preferring an existing path whenever it fits. Do not add, delete, split, merge, or rewrite candidates. Do not alter Claim, confidence, or evidence. Return one decision for every input index.
\end{tcolorbox}

\subsection{Maintenance}
\label{app:method-maintenance}
\textbf{T1/T2} use deterministic updates. A repeated value takes the higher of the existing and incoming confidence scores and advances confirmation; a conflicting value replaces it only with strictly higher confidence. An existing value at the confidence ceiling of 5 is therefore retained under this rule. Coupled slots compare their strongest confidence and retain unmentioned values only if consistent. Deterministic constraints remove invalid existence/count, partner, life-stage, and occupation combinations.

\textbf{T3/T4} operate within one path: \textsc{Keep} preserves, \textsc{Add} accepts a distinct Claim, \textsc{Update} revises the same characteristic, and \textsc{Retire} requires direct contradiction. The abridged instruction accounts for each input exactly once and forbids cross-path merging; evidence endpoints are recomputed from validated sources. Invalid responses fall back to keeping old and appending new Claims.

Unobserved continuous paths remain Active; non-continuous paths may become Dormant and reactivate with renewed support. Sustained absence can trigger T3 continuity review. The availability instruction treats absence as insufficient counter-evidence and defaults to keeping a path Active when uncertain.

The abridged maintenance prompts below are translated from Chinese and apply only to T3/T4.

\begin{tcolorbox}[breakable, colback=black!2, colframe=black!25, boxrule=0.3pt, arc=1mm, left=4pt, right=4pt, top=3pt, bottom=3pt, before skip=4pt, after skip=4pt]
\footnotesize\ttfamily\RaggedRight
        \#\# T3/T4 Claim maintenance\\
        Input: \{previous Claims\}, \{current Claims\}, grouped by path\\
        Process each path independently. KEEP or ADD distinct characteristics, UPDATE only the same characteristic, and RETIRE only under direct contradiction. Account for every input Claim exactly once; never merge across paths or output evidence.\\[0.3em]
        \#\# Continuity\\
        Label a new path continuous if it should remain valid without repeated observation; label current needs and finite tasks non-continuous.\\[0.3em]
        \#\# Dormancy\\
        For an unobserved non-continuous path, choose KEEP\_ACTIVE or MOVE\_DORMANT. Absence alone is not counter-evidence; uncertainty defaults to KEEP\_ACTIVE.
\end{tcolorbox}

\subsection{Readout}
\label{app:method-readout}
F5 deterministically emits Active Claim text, confidence, and stable/stage-specific scope. Separately, it ranks ledger-derived long-term aggregates and recent item activity by engagement and recency. These explicit facts and inferred Claims share a query-independent readout; F5 calls no model.

\section{BehaviorMemBench Experimental Details}
\label{app:exp}
\label{app:exp-behaviormembench}

\paragraph{\textbf{Readout ablation results.}}
\label{app:bmb-readout-ablation}
Table~\ref{tab:ablation-main} gives all nine metrics for Figure~\ref{fig:readout-ablations}; all masks reuse one frozen Full state with a Qwen3-32B builder and reader.

\begin{table*}[t]
    \centering
    \caption{BehaviorMemBench readout ablations with a Qwen3-32B builder and task reader.}
    \label{tab:ablation-main}
    \footnotesize
    \setlength{\tabcolsep}{2.4pt}
    \renewcommand{\arraystretch}{1.06}
    \begin{tabular*}{\textwidth}{@{\extracolsep{\fill}}lcccc@{\hspace{7pt}}ccccc@{}}
        \toprule
        & \multicolumn{4}{c}{\cellcolor{TableHeader}Remember pool} & \multicolumn{5}{c}{\cellcolor{TableHeader}Infer pool} \\
        \cmidrule(lr){2-5}\cmidrule(lr){6-10}
        & Recall $\uparrow$ & \multicolumn{3}{c}{Exposure $\downarrow$} & \multicolumn{3}{c}{Recall $\uparrow$} & \multicolumn{2}{c}{Exposure $\downarrow$} \\
        \cmidrule(lr){2-2}\cmidrule(lr){3-5}\cmidrule(lr){6-8}\cmidrule(lr){9-10}
        Configuration & $R$ & $E_{\mathrm{cluster}}$ & $E_{\mathrm{history}}$ & $E_{\mathrm{random}}$ & $R$ & $R_{\mathrm{related}}$ & $R_{\mathrm{new}}$ & $E_{\mathrm{history}}$ & $E_{\mathrm{random}}$ \\
        \midrule
        \TableRowShade{TableHighlight}{\textwidth}
        Full BehaviorMem & \textbf{0.612} & \textbf{0.111} & 0.221 & \underline{0.025} & \textbf{0.443} & \textbf{0.461} & \underline{0.416} & \underline{0.226} & \underline{0.161} \\
        w/o Explicit Memory & 0.529 & 0.156 & \textbf{0.205} & 0.042 & 0.438 & \underline{0.454} & 0.410 & 0.229 & 0.163 \\
        w/o Implicit Memory & 0.602 & 0.116 & 0.222 & \textbf{0.018} & 0.432 & 0.443 & 0.409 & 0.227 & 0.165 \\
        w/o T1/T2 Claims & \underline{0.608} & 0.114 & \underline{0.216} & 0.027 & 0.435 & 0.450 & 0.409 & \textbf{0.225} & 0.165 \\
        w/o T3/T4 Claims & 0.596 & \underline{0.113} & 0.237 & \underline{0.025} & \underline{0.442} & 0.449 & \textbf{0.423} & 0.228 & \textbf{0.159} \\
        \bottomrule
    \end{tabular*}
\end{table*}

\subsection{Dataset and Model Settings}
\label{app:bmb-dataset}
Data and reader rules follow Appendix~\ref{app:bench}; model assignments follow Section~\ref{sec:experimental-setup}. Activity groups keep batch-count ties together: 1--6, 7--11, 12--16, 17--21, and at least 22 batches. Within each builder and pool, activity analysis uses transitions common to five agentic methods and user macro-averaging; target-familiarity analysis likewise uses common valid transitions within each configuration.

\subsection{Baselines and Adaptation}
\label{app:bmb-baselines}
Methods process histories chronologically into native query-independent representations, without candidate access during construction.

\textbf{Long Context} renders dated visits with titles, categories, prices, and engagement, retaining recent lines within 8{,}000 tokens. \textbf{Frequency Profile} ranks the top 30 L2 categories using click/browse/favorite-or-cart/purchase weights 1/2/3/5 and $0.9^k$ decay for visits $k$ batches earlier. \textbf{Template Profile} uses the same weights in fixed category, price, engagement, and title-fragment slots; new categories first occur within three batches, and inactive categories are absent for at least four.

For \textbf{A-Mem}, \textbf{Mem0}, and \textbf{LightMem}, one batch becomes one user message. Readout enumerates recent stored entries within 8{,}000 tokens. A-Mem exposes generated context and keywords, excluding raw notes. Mem0 lists current facts by creation time. LightMem disables pre-compression and topic segmentation and forces extraction each batch, evaluating this configured pipeline rather than its full efficiency-oriented setup.

\textbf{Persona Tree} applies add/update/delete/no-op operations to its fixed hierarchy. The complete schema remains available during construction; only readout prunes empty leaves and uses an empty description placeholder. We evaluate state construction without MemListener training, query-dependent routing, or retrieval fusion.

\section{MemoryCD Experimental Details}
\label{app:exp-memorycd}

\begin{table*}[t]
    \centering
    \caption{MemoryCD results under a Qwen3-32B task reader, with both memory builders.}
    \label{tab:memorycd-reader-q32}
    \footnotesize
    \setlength{\tabcolsep}{2.0pt}
    \renewcommand{\arraystretch}{0.98}
    \begin{tabular*}{0.96\textwidth}{@{\extracolsep{\fill}}lcccc@{\hspace{6pt}}cccc@{}}
        \toprule
        \rowcolor{TableHeader}
        Memory & \multicolumn{4}{c}{\textbf{Single-domain}} & \multicolumn{4}{c}{\textbf{Cross-domain}} \\
        \cmidrule(lr){2-5}\cmidrule(lr){6-9}
        & HR@10 $\uparrow$ & N@10 $\uparrow$ & MAE $\downarrow$ & RMSE $\downarrow$ & HR@10 $\uparrow$ & N@10 $\uparrow$ & MAE $\downarrow$ & RMSE $\downarrow$ \\
        \midrule
        \rowcolor{TableSection}
        \multicolumn{9}{l}{\emph{Non-parametric memories}} \\
        Long Context & \underline{0.615} & \underline{0.313} & \textbf{0.444} & \textbf{0.821} & 0.589 & 0.268 & \textbf{0.447} & \textbf{0.783} \\
        Frequency Profile & 0.607 & 0.289 & 0.812 & 1.046 & \underline{0.623} & \underline{0.304} & 0.958 & 1.164 \\
        Template Profile & \textbf{0.639} & \textbf{0.320} & \underline{0.758} & \underline{0.998} & \textbf{0.661} & \textbf{0.336} & \underline{0.901} & \underline{1.111} \\
        \cmidrule(lr){1-9}
        \rowcolor{TableSection}
        \multicolumn{9}{l}{\emph{Builder: Qwen3-32B}} \\
        Mem0 & 0.637 & 0.320 & 0.613 & 0.972 & 0.596 & 0.275 & 0.817 & 1.178 \\
        A-Mem & \textbf{0.712} & \textbf{0.386} & 0.676 & 1.029 & \textbf{0.665} & \textbf{0.324} & 0.930 & 1.183 \\
        Persona Tree & 0.684 & 0.366 & 0.626 & 0.957 & \underline{0.661} & 0.315 & 0.704 & 0.970 \\
        LightMem & 0.674 & 0.371 & \textbf{0.487} & \textbf{0.863} & 0.635 & 0.309 & \textbf{0.499} & \textbf{0.830} \\
        \TableRowShade{TableHighlight}{0.96\textwidth}
        \textbf{BehaviorMem} & \underline{0.700} & \underline{0.372} & \underline{0.581} & \underline{0.918} & \underline{0.661} & \underline{0.317} & \underline{0.598} & \underline{0.897} \\
        \cmidrule(lr){1-9}
        \rowcolor{TableSection}
        \multicolumn{9}{l}{\emph{Builder: Qwen3.6-35B-A3B}} \\
        Mem0 & 0.648 & 0.325 & 0.674 & 1.034 & 0.614 & 0.281 & 0.942 & 1.280 \\
        A-Mem & 0.694 & \underline{0.375} & 0.688 & 1.028 & \underline{0.673} & \underline{0.324} & 0.881 & 1.126 \\
        Persona Tree & \underline{0.700} & 0.373 & 0.656 & 0.971 & 0.636 & 0.308 & \underline{0.652} & \underline{0.913} \\
        LightMem & 0.670 & 0.368 & \textbf{0.549} & \textbf{0.910} & 0.657 & 0.319 & \textbf{0.598} & \textbf{0.886} \\
        \TableRowShade{TableHighlight}{0.96\textwidth}
        \textbf{BehaviorMem} & \textbf{0.710} & \textbf{0.388} & \underline{0.638} & \underline{0.949} & \textbf{0.682} & \textbf{0.345} & 0.720 & 0.988 \\
        \bottomrule
    \end{tabular*}
\end{table*}

\subsection{Dataset and Tasks}
\label{app:memorycd-dataset}
MemoryCD~\citep{zhang2026memorycd} uses Amazon Reviews 2023 product-review histories containing item information, review text, and observed ratings. We consider Beauty, Books, Electronics, and Home and Kitchen. In the \textbf{single-domain} setting, memory uses the target-domain history preceding the final three interactions, which are reserved for testing. In the \textbf{cross-domain} setting, memory combines interactions from Beauty, Books, and Electronics, and testing uses the final three Home and Kitchen interactions for a 323-user cohort. 
\subsection{Baselines and Implementation}
\label{app:memorycd-implementation}
Methods use the adapters in Appendix~\ref{app:bmb-baselines} under a shared readout token budget. Agentic builders and task readers use Qwen3-32B or Qwen3.6-35B-A3B; non-parametric baselines need no LLM builder.

\subsection{Evaluation Protocol}
\label{app:memorycd-evaluation}
\label{app:memorycd-results}
For \textbf{item ranking}, the reader ranks benchmark-provided candidates containing one held-out positive and negatives. HR@10 measures whether the positive ranks in the top ten; NDCG@10 also rewards higher placement. For \textbf{rating prediction}, the reader predicts a score on the 1--5 scale; MAE measures average absolute error, and RMSE weights larger errors more heavily. Single-domain results average metrics equally across four domains. Tables~\ref{tab:memorycd-builders} and~\ref{tab:memorycd-reader-q32} report both readers, with N@10 denoting NDCG@10. 

\end{document}